\documentclass[letterpaper, 10 pt, conference]{ieeeconf}  

\IEEEoverridecommandlockouts                              

\usepackage{stfloats}
\usepackage{float}
\usepackage{enumerate}
\usepackage{amssymb}
\usepackage{bm}
\usepackage{amsmath}
\usepackage{geometry}
\usepackage{multirow}
\usepackage{threeparttable}
\usepackage{graphicx}
\usepackage{hyperref}
\usepackage{algorithm}
\usepackage{algorithmic}

\title{\LARGE \bf
Learning-Based Framework for Camera Calibration with Distortion Correction and High Precision Feature Detection
}

\author{Yesheng Zhang$^{1,3}$ , Xu Zhao$^{1,2}$ and Dahong Qian$^{1,3}$ 
\thanks{$^{1}$ Institute of Medical Robotics, Shanghai Jiao Tong University, Shanghai 200240, China}
\thanks{$^{2}$ The Department of Automation, Shanghai Jiao Tong University, Shanghai 200240, China}
\thanks{$^{3}$ School of Biomedical Engineering, Shanghai Jiao Tong University, Shanghai 200240, China}
}

\begin{document}

\maketitle
\thispagestyle{empty}
\pagestyle{empty}

\begin{abstract}
Camera calibration is a crucial technique which significantly influences the performance of many robotic systems.
Robustness and high precision have always been the pursuit of diverse calibration methods.
State-of-the-art calibration techniques, however, still suffer from inexact corner detection, radial lens distortion and unstable parameter estimation.
Therefore, in this paper, we propose a hybrid camera calibration framework which combines learning-based approaches with traditional methods to handle these bottlenecks. In particular, this framework first leverages CNN to perform efficient distortion correction. Then, sub-pixel chessboard corner detection are achieved by heatmap learning with specially-designed post-processing.
To increase the stability of parameter estimation, an image-level RANSAC-based calibration is integrated in this framework.
Compared with state-of-art methods, experiment results on both real and synthetic datasets manifest the better robustness and higher precision of the proposed framework.
The massive synthetic dataset is the basis of our framework's decent performance and will be publicly available along with the code at \href{https://github.com/Easonyesheng/CCS}{https://github.com/Easonyesheng/CCS}.
\end{abstract}

\section{INTRODUCTION}
Camera calibration is crucial for many robotic applications\cite{engel2017direct, SLAM:survey, CalibtoAD}. 
Especially, in some industrial and medical applications \cite{barreto2009automatic}, precision and robustness of camera calibration have a significant impact on the overall performance.


   \begin{figure*}[thpb]
      \centering
      \includegraphics[width=\textwidth]{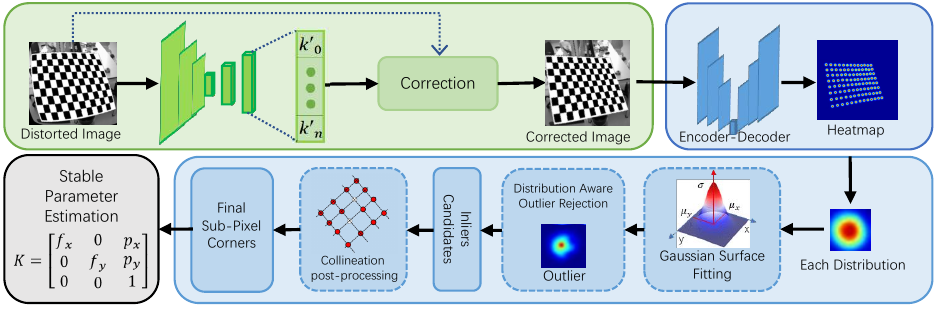}
      \caption{Our framework includes three main parts. First, radial distortion is removed by correction model inferred from distorted images by our network (green part). The chessboard corner detection part consists of learning heatmap and sub-pixel refinement with outlier rejection and collineation post-processing (blue part). After precise sub-pixel corner coordinates are obtained, stable parameter estimation owning to a simplified RANSAC procedure is performed to achieve accurate camera parameters (gray part). }
      \label{figMain}
   \end{figure*}

The most widely-used camera calibration toolboxes \cite{opencv_library,MATLAB:2010} are built based on Zhang's technique \cite{Zhang:2000}. 
Usually chessboard images are captured to calculate camera parameters according to the established feature correspondences between 3D world and 2D images. This pipeline is flexible and easy to implement. Building a precise and robust calibration system, however, is still a challenging problem, mainly due to the following issues:
\begin{enumerate}[1)]
\item \textbf{Inexact detection}. Sub-pixel feature localization is hard to achieve, especially in scenario with noise and bad illumination.
\item \textbf{Radial distortion}. Severe radial lens distortion may result in calibration failure.
\item \textbf{Unstable estimation}. Purely algebraic optimization of re-projection error leads to sub-optimal and unstable calibration results. 
\end{enumerate}

\begin{figure}[!t]
\centering
\includegraphics[width=\linewidth]{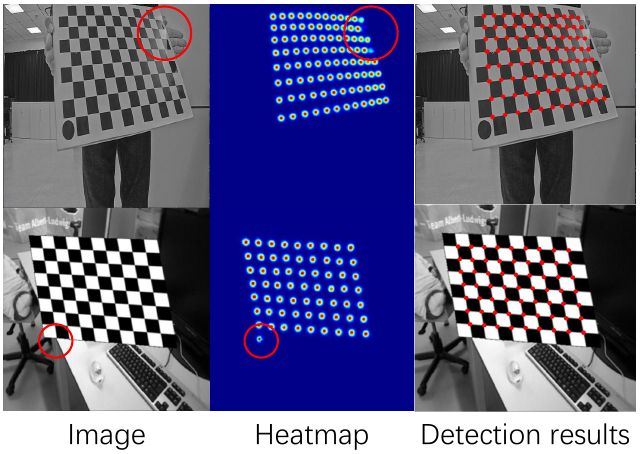}
\caption{Examples of detection outliers and detection results. The real image (up) and its native heatmap are related to lost corners and the synthetic ones (bottom) are fake corners. Both of them can be rejected according to abnormal distribution and achieve accurate detection.}
\label{fig_outlier}
\end{figure}

For the first issue, many early works \cite{Lucchese:2002,SimonPlacht:2014ty,Geiger12} make efforts to sub-pixel feature localization using hand-crafted features, but are sensitive to noises. 
As the rapid advance of deep learning, convolution neural network (CNN) is introduced \cite{Donne:2016eq,BenChen:2018fy,Wu:2021} to detect chessboard corners. 
Although they are more robust but still limited in pixel level precision.
Recently, Chen et al. \cite{chen2021automatic} propose a chessboard corner detection framework which refines the peak of CNN's response map to get the sub-pixel corner location.
This method achieves promising results, but its detection network is trained with corner coordinates to generate response map with unknown distribution, which is not closely consistent with its corner refinement.
In contrast, we propose another way to better combine the robustness of CNN detection and the precision of sub-pixel refinement.
We design a heatmap where each corner is represented by a Gaussian distribution and its center is the labeled sub-pixel location. 
Then we train our detection network with ground truth heatmaps and apply Gaussian surface fitting algorithm to get sub-pixel location from the network's output.
This method achieves accurate detection, but we found the unreliable detection with abnormal distribution occurs in practice (Fig. \ref{fig_outlier}).
Thus, we propose the distribution-aware outlier rejection to eliminate unreliable detection.
To fix the corners after outlier rejection and achieve better precision, the collineation post-processing is applied to achieve final corners.

The second issue is about lens distortion, which is a non-trivial problem in camera calibration, especially radial distortion\cite{Tang:2020bo}.
Classical methods \cite{salvi1998approach,Tsai:1987,Zhang:2000} estimate camera and distortion parameters simultaneously by iterative optimization.
However, ambiguity is introduced in this way as parameters are tangled together leading to failure under severe distortion. 
On the other hand, the distortion is related to the curvature of the straight lines.
This relationship can be easily learned by deep network and utilized to correct distortion \cite{Xue_2019_CVPR,Yin:2018tc,Zhao:jp}, and chessboard images naturally own many straight lines.
Thus we apply CNN to infer the correction parameters from distorted images with a practical distortion model. As the corner detection accuracy suffers from radial distortion and our collineation post-processing assumes no distortion, so the distortion correction is taken as the first step in the proposed framework.

The third issue is caused by purely algebraic optimization which may lead to unreasonable calibration results. Some works \cite{Sagawa:2008, GeoCalib} propose geometry-based algorithms to address this issue, but not complete enough to be widely used. Random Sample Consensus (RANSAC) \cite{RANSAC} algorithm has been introduced into calibration to eliminate outliers to enhance the stability of parameter estimation \cite{ZHOU_ransac_points, LV_ransac_pose}. We adopt a simplified RANSAC-based calibration procedure to improve the robustness of parameter estimation. Different from previous work, as outlier rejection already gets involved in corner detection, we propose an efficient image-level RANSAC algorithm based on reprojection error to search for the optimal camera model.

In sum, the critical components of a calibration system, distortion correction, feature detection and parameter estimation, are reforged and integrated as a novel and efficient calibration framework (Fig. \ref{figMain}). The contributions can be summarized as follows.
\begin{enumerate}
    \item A novel camera calibration framework is proposed, which includes radial distortion correction, sub-pixel feature detection and stable parameter estimation.
    \item We design a learning heatmap method with outlier rejection and collineation post-processing to achieve sub-pixel chessboard corner detection.
    \item The new framework achieves precise results on both synthetic and real data, outperforming state-of-art methods by a noteworthy margin.
\end{enumerate}
\section{Related work}\label{rw}

\textbf{Corner Detection.} 
The widely-used OpenCV\cite{opencv_library} detection function refines coordinates according to gray distribution constraints.
In \cite{Geiger12}, corner coordinates are refined based on the chessboard structure.
While these methods achieve sub-pixel precision, heavily relying on hand-crafted features leads to the lack of robustness.
Recently, some detection algorithms based on learning features are proposed \cite{Donne:2016eq,BenChen:2018fy,Wu:2021}. 
They are robust against noise but trapped in pixel level accuracy.
Schroter et al. \cite{schroeter2020learning} propose a learning-based general point detection method.
This method achieves sub-pixel accuracy, yet provides false negative results owning to global detection.
Kang et al. \cite{kang2021sparse} tackle this problem by parsing global context, but perform weak in difficult detection environment.
Chen et al. \cite{chen2021automatic} extend learning-based chessboard corner detection to sub-pixel accuracy by fitting CNN's response map.
This pipeline is close to ours, but our learning heatmap method gets more tight connection between detection and refinement \cite{zhang2020distribution}.
We also propose some specific techniques considering the peculiarity of chessboard pattern. 

\textbf{Radial Distortion Correction.} Classical algorithms integrate distortion in camera model and solve it by non-linear optimization techniques \cite{salvi1998approach, Tsai:1987, Zhang:2000}
Although this method works well under slight radial distortion, they may end up with a bad solution when the distortion is severe.
On the other hand, straight lines wrapped by distortion are appropriate features for CNNs to learn distortion parameters \cite{Rong:wf,Xue_2019_CVPR,Yin:2018tc}.
As chessboard images contain sufficient straight line features, we adopt deep CNNs to regress parameters of a specially selected distortion correction model.


\textbf{Parameter Estimation.} Widely-used Zhang's technique \cite{Zhang:2000} first solves the initial guess of parameters based on the correspondence between the real world and image.
Then these parameters are refined by minimizing the reprojection error.
However, purely algebraic optimization is unstable leading to suboptimal calibration results.
To address this issue, some RANSAC-based algorithms \cite{LV_ransac_pose,ZHOU_ransac_points} are proposed to achieve robust parameter estimation by excluding unreliable images or corners.

\section{The proposed calibration Framework}\label{sec_sys}
In this section, we introduce our camera calibration framework by describing each component.
\subsection{Radial Distortion Correction}
As simultaneous parameter estimation and distortion correction not only increase calibration effort but also reduce precision, our framework performs distortion correction first.
Considering that chessboard images contain sufficient straight line features which are helpful for CNN to learn the distortion model \cite{Xue_2019_CVPR,Zhao:jp}, we adopt a CNN-based, 8 layers encoder with 3 regression layers to regress correction model parameters from images.
Besides, we apply a more practical distortion model than previous work: the radial model ($r_d = r_c(k_0 + k_1r_c^2 + \dots) $), which is symmetric and flexible \cite{Tang:2020bo}.
Its symmetric property maintains consistency between the distortion and correction.
Thus we can generate distorted images to train this network, which outputs parameters of another radial model with higher order ($r_c = r_d(k_0^{'}+k_1^{'}r_d+k_2^{'}r_d^2+\dots)$).
Its flexibility allows us to choose appropriate number of output parameters according to the specific distortion.
In practice, to train the network, we generate massive distorted chessboard image by randomly sampling 3 parameters in distortion model and the output parameter number is set to 5.
Some image examples can be seen in Fig. \ref{fig_ex_d}.
As the distortion model and correction model have different parameters, only robust sampling grid loss is adopted \cite{Zhao:jp}:
\begin{equation}
    L_{grid} =\frac{1}{N} \sum_i^N ||p_{dst}^i - p_{cor}^i||_1
\end{equation}
where $p_{dst}$ represent the location of distorted grid points and $p_{cor}$ represent the corrected ones.

\subsection{Chessboard Corner Detection}
We propose the learning heatmap refinement method to combine the robustness of network detection with the precision of sub-pixel refinement.

\textbf{Learning heatmap detection.} The ground truth heatmap ($Y$) is designed to represent each corner as a 2-dimensional Gaussian distribution ($\mathcal{G}$) centered at the labelled sub-pixel coordinate.
Than we use these heatmaps to train the classical UNet \cite{unet} with detection loss:
\begin{equation}
    L_{detect} = \iint_{R^2} \|\hat{Y}(x,y) - Y(x,y) \|^2 dxdy
\end{equation}
where $\hat{Y}$ is the network output.

\textbf{Sub-pixel refinement with outlier rejection.}
After the network transforms chessboard image into a heatmap ($\hat{Y}$), we apply Gaussian surface fitting algorithm on each distribution ($\hat{\mathcal{G}}$) in the heatmap.
Specifically, for each distribution, we find the center $\bm{\mu}$ and the variance $\bm{\sigma}$ as follow: 
\begin{equation}
    \arg \min_{\bm{\mu}, \bm{\sigma}} \left \|\mathcal{
    G}( \bm{\mu},\bm{\sigma}) - \hat{\mathcal{G}}\right \|^2 _2
\end{equation}
which can be solved by the SVD decomposition using points ($I_i(x,y)$) around the distribution peak:
\begin{equation}
    I_i(x,y) = e^{-\frac{(x_i-\mu_x)^2}{2\sigma_x^2} - \frac{(y_i-\mu_y)^2}{2\sigma_y^2}}, 
\end{equation}
Then we have:
\begin{small}
\begin{equation*}
	\begin{matrix}
	I_i \times \ln{I_i} 
	= 
		\left[
            \begin{array}{ccccc}
                 I_i \! & I_i x_i \!& I_i y_i \!& I_i x_i^2 \!& I_i y_i^2
            \end{array}
        \right]
		\cdot
		\\ \left[
            \begin{array}{ccccc}
                 -\frac{\mu_x^2}{2\sigma_x^2}-\frac{\mu_y^2}{2\sigma_y^2} &
                 \frac{\mu_x}{\sigma_x^2} &
                 \frac{\mu_y}{\sigma_y^2} &
                 - \frac{1}{2\sigma_x^2} &
                 - \frac{1}{2\sigma_y^2}
            \end{array}
        \!\right]^T
	\end{matrix}
\end{equation*}	
\end{small}
which can be expressed as:
\begin{equation}
    a_i = b_i \cdot c_i^T
\end{equation}
For N points, we can stack formulations as:
\begin{equation}
    A = BC^T
\end{equation}
Then we can solve the matrix $C^T$ which contains $\bm{\mu}$ and $\bm{\sigma}$ by the SVD decomposition.
The $\bm{\mu} = (\mu_x,\mu_y)$ represents the corner's sub-pixel coordinate.
As our network is trained to generate standard Gaussian distribution, abnormal distribution corresponds to unreliable detection.
The distribution aware outlier rejection is to eliminate wrong detection according to $\bm{\sigma}$ compared with the variance of training data.

\begin{figure}[!h]
\centering
\includegraphics[width=\linewidth]{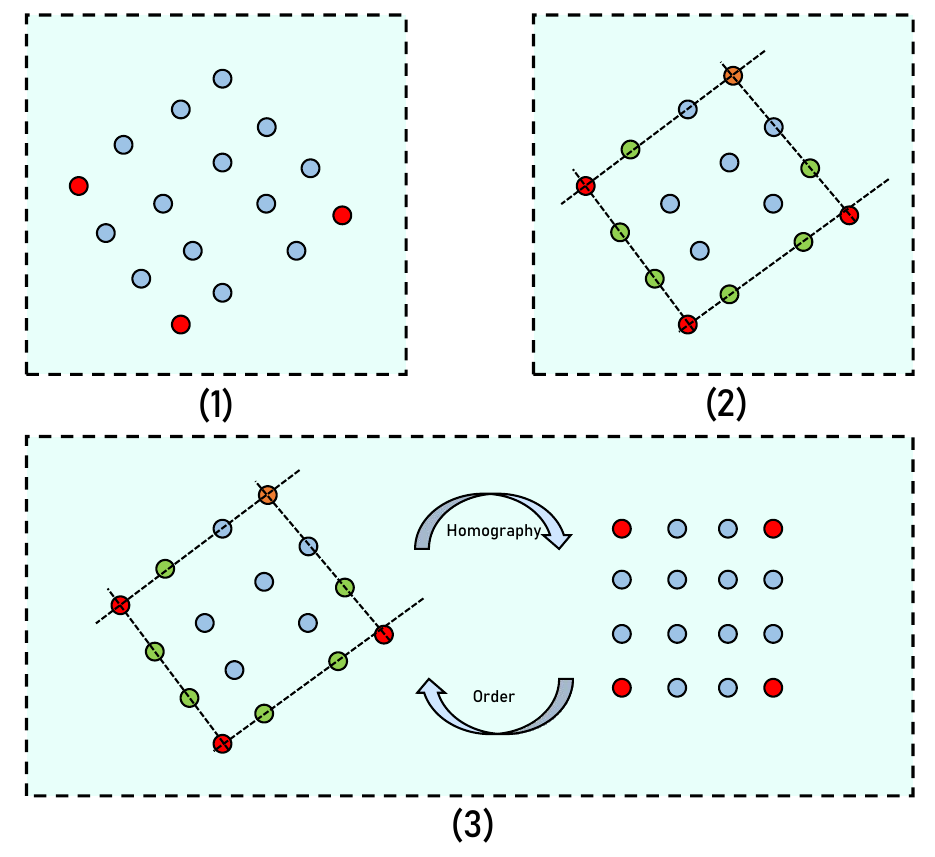}
\caption{The corner sorting process.}
\label{fig_sort}
\end{figure}

\textbf{Collineation post-processing}
After inliers are selected, collineation post-processing is proposed not only to refine the sub-pixel coordinates but also to recover some lost corners.
However, before collineation post-processing, we need to sort unordered corners after outlier rejection.
Therefore we can get sets of corners belong to each line.
The sort method is described as follow (Fig. \ref{fig_sort}).
\begin{enumerate}[(1)]
    \item Find the extreme points (green points) by coordinates.
    \item Get the chessboard edge (red lines) through extreme points and their neighbors, and intersect lines to get the last extreme point (yellow point).
    \item Calculate the homography transformation from extreme points to a standard rectangle's four endpoints, then the original points are indexed in the order of the transformed points.
\end{enumerate}
Due to the distortion being removed firstly by our framework, collineation post-processing utilizes sorted inlier corners to fit lines and then calculate the final coordinates by intersecting these lines.

\subsection{Parameter Estimation}
In order to achieve robust parameter estimation, we propose a simple yet effective parameter estimation method which improves Zhang's technique \cite{Zhang:2000} by simplified RANSAC.
Different with early works \cite{LV_ransac_pose, ZHOU_ransac_points}, the proposed method is aimed at searching for the best camera model which keeps consistency in reprojection error at image level, because our detection part already eliminates outlier points.
This method can be illuminated as follows:
\begin{enumerate}
    \item Choose some of the images randomly to estimate parameters based on Zhang's technique.
    \item Calculate the reprojection error of all images and determine inliers whose reprojection errors are less than the threshold.
    \item Output the parameters if the inliers number is big enough; otherwise repeat the above steps.
\end{enumerate}

\section{Experimental Results}\label{sec_exp}
The performance of the proposed camera calibration framework is evaluated on both synthetic and real data.
We also construct experiments to demonstrate the capabilities of our distortion correction part and corner detection part respectively.
\subsection{Dataset and Metric}
\begin{figure}[!t]
\centering
\includegraphics[width=\linewidth,scale=0.5]{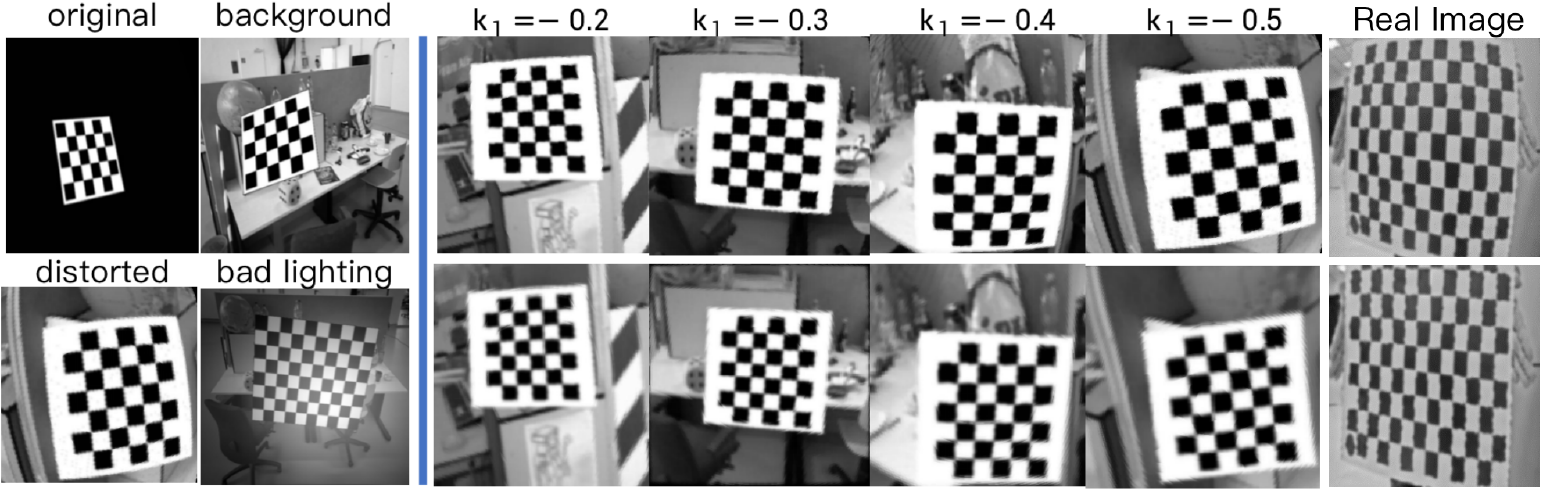}
\caption{\textbf{Left:} Examples of synthetic images. \textbf{Right:} Examples of images under different degrees of distortion($k_0 = 1, -0.5\leq k_1 \leq -0.2, k_2 = -0.02$) (top) along with the corrected images (bottom) acquired by our method.}
\label{fig_ex_d}
\end{figure}

To train our networks, we generate massive chessboard images ($480 \times 480$) with ground truth corner heatmaps and camera parameters (focal length, principal points and scale) in pixel ($100 \leq f_x,f_y \leq 300; 120 \leq p_x,p_y \leq 360; 1 \leq s \leq 5$).
Moreover, noise, bad lighting, distortion and fake background using TUM dataset \cite{TUMdata} are applied as data augmentation.
Specifically, the image distortion level is decided by parameters $k_0, k_1$ and $k_2$.
The example synthetic images can be seen in Fig. \ref{fig_ex_d} (left). 
We use the metrics related to focal length (FL) and principal points (PP) in intrinsic matrix which can be defined as:
\begin{equation}\small
    E_{FL} = \Vert FL_{GT} - \hat{FL} \Vert_1
\end{equation}
\begin{equation}\small
    E_{PP} = \Vert PP_{GT} - \hat{PP} \Vert_1
\end{equation}
\begin{equation}\small
    E_{IP} =  \frac{1}{2} (E_{FL} + E_{PP})
\end{equation}

Classical reprojection error (RPE) is used as well.
\begin{equation}
   RPE = \frac{1}{N\times M}\sum_i^N \sum_j^M \|p_{ij} - sK[R_i|t_i]P_{ij}\|_2^2
\end{equation}
where $p_{ij}$ are the image points and $P_{ij}$ are corresponding world points. $M$ is the number of chessboard corners. $N$ is the number of images.


\begin{figure}[!t]
\centering
\includegraphics[width=\linewidth]{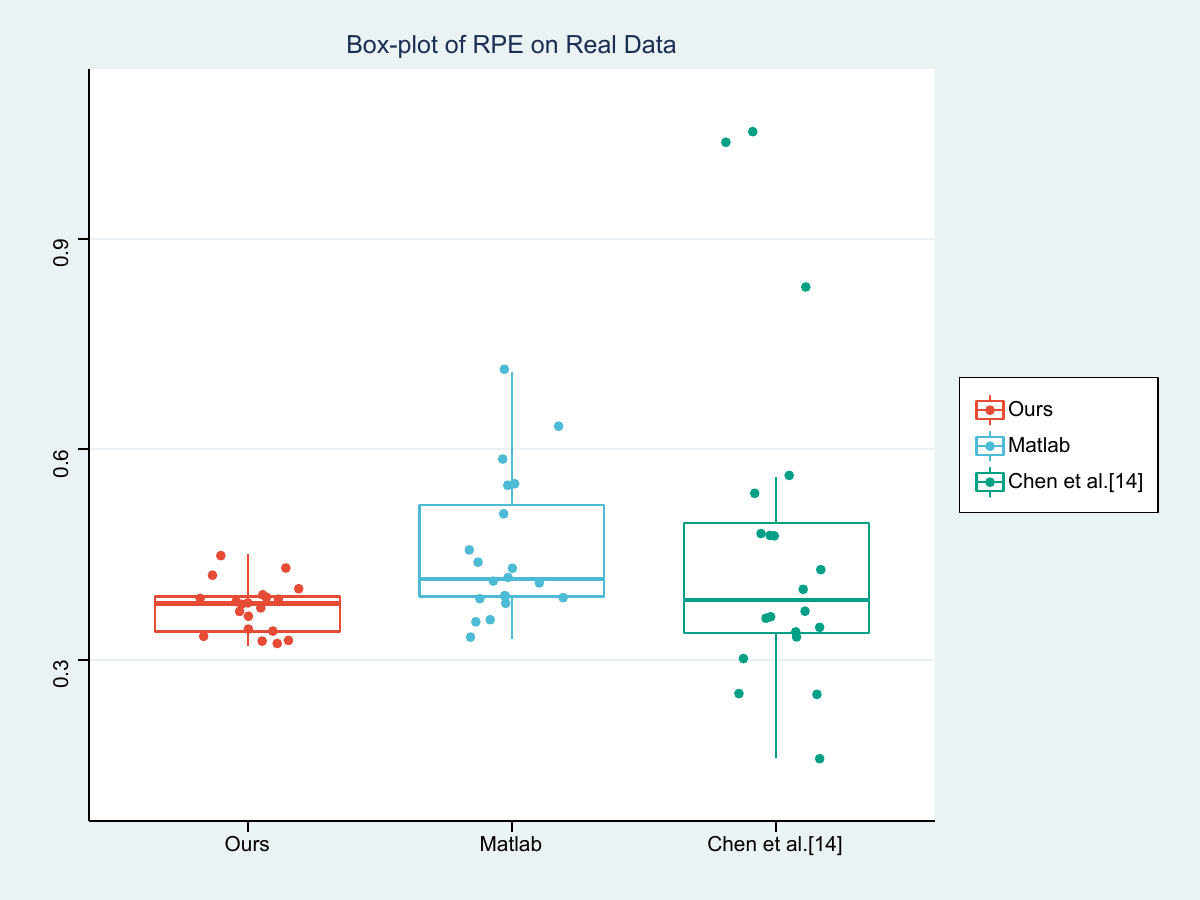}
\caption{The box-plot of calibration RPE on real data.}
\label{fig_ipe_d}
\end{figure}

\subsection{Calibration Performance}
To extensively evaluate our framework, we conduct calibration experiments on four different image configurations: noise, bad lighting, distortion and real data.
Our framework is compared with two state-of-art calibration methods.
\textbf{1)} The \textbf{Matlab calibration toolbox} \cite{MATLAB:2010} represents traditional methods, which contains complicated process for precision and robustness.
\textbf{2)} The method proposed by \textbf{Chen et al.} \cite{chen2021automatic} which performs learning-based chessboard corner detection and it achieves sub-pixel precision and more accurate than kang et al. \cite{kang2021sparse} in detection experiment (Table. \ref{table_cl} ).
We train its network on our synthetic dataset and combine it with OpenCV \cite{opencv_library} to perform distortion correction and parameter estimation.


\textbf{Calibration under noise and bad lighting.} 
The first experiment is to demonstrate the robustness and accuracy of our framework in terms of environmental noise like low sensor resolution and uneven illumination.
We apply $3 \times 3$ Gaussian kernel with $\sigma=1.5$ to blur images for noise simulation.
The uneven brightness is simulated by specular lighting model \cite{phong1975illumination} with random size and center.
The average results of 50 independent trials are summarized in the first and second rows of Table \ref{table_noise}.
It can be seen that our framework outperforms the other methods in general.
The Matlab method's error concentrates on focal length.
Chen et al. method achieves promise accuracy but its variance is high indicating the instability. 
Besides, the low variance manifest the stability of our framework owing to the robust parameter estimation.

\begin{table*}[!t]
\small
\begin{threeparttable}[b]
\caption{Comparison of camera calibration with different images.}
\centering
\label{table_noise}
\setlength{\tabcolsep}{1mm}{
\begin{tabular}{|c|c|c|c|c|c||c|c|c|c|c||c|c|c|c|c||}
\hline
 \multirow{3}{*}{Images}  & \multicolumn{15}{c|}{Calibration errors(pixels) }                               \\ \cline{2-16} 
          &                                          \multicolumn{5}{c|}{Ours}   & \multicolumn{5}{c|}{ Matlab} & \multicolumn{5}{c|}{Chen et al. \cite{chen2021automatic}} \\ \cline{2-16} 
                          & $E_{FL}$   & $E_{PP}$   & $E_{IP}$  & $RPE$ & $\sigma^2$\tnote{*}  & $E_{FL}$   & $E_{PP}$     & $E_{IP}$ & $RPE$  & $\sigma^2$\tnote{*}  & $E_{FL}$   & $E_{PP}$  & $E_{IP}$ & $RPE$ & $\sigma^2$\tnote{*} \\ \hline
 noise       & \textbf{0.42}    & 0.94 & \textbf{0.68} & \textbf{0.23} & 0.22 & 2.41  & \textbf{0.63}  & 1.52 & 0.53 & 0.73 & 0.51  & 1.08  & 0.79 & 0.46 & 0.69   \\ \hline 
 bad lighting         & \textbf{0.84}    & \textbf{0.64} & \textbf{0.74} & 0.39 & 0.27 & 2.38  & 1.48  & 1.93 & 0.47 & 0.79 & 0.88 & 0.73  & 0.81 & \textbf{0.35} & 0.78 \\ \hline
 distortion level 1            & \textbf{0.78}    & \textbf{0.41} & \textbf{0.60} & \textbf{0.55} & 0.34 & 3.25  & 2.16  & 2.71 & 0.64 & 0.77 & 0.98 & 1.34  & 1.16 & 1.02 & 0.73 \\ \hline
 distortion level 2           & \textbf{0.98}    & \textbf{1.26} & \textbf{1.12} & \textbf{0.53} & 0.42 & 4.35  & 6.33  & 5.35 & 0.57 & 0.87 & 2.13 & 3.25  & 2.69 & 0.97 & 0.83\\ \hline
\end{tabular}}
\begin{tablenotes}
    \footnotesize
    \item[*] The variance of $E_{IP}$.
\end{tablenotes}
\end{threeparttable}
\end{table*}

\textbf{Calibration under distortion.} 
We also conduct calibration experiments on distorted images, and we set two different distortion levels by randomly setting 3 parameters.
The first distortion level's parameters are: $k_0 = 1, -0.2 \leq k_1 \leq -0.35, -0.1 \leq k_3 \leq 0 $.
The second distortion level is more severe, whose parameters are: $0.8 \leq k_0 \leq 1.2, -0.35 \leq k_1 \leq -0.5, -0.3 \leq k_3 \leq -0.1 $.
For each distortion level, 50 independent trials of calibration are performed and average results are shown in the last two rows of Table. \ref{table_noise}.
Our method maintains high precision under different distortion levels, but other methods get accuracy reduce due to their inefficient non-linear distortion correction method.
Especially under severe distortion, the best precision demonstrates the robustness of our method against distortion.

\begin{table}[!t]
\small
\caption{Comparison of calibration on real data.}
\centering
\label{table_rd}
\begin{tabular}{|c||c||c||c|}
\hline
    & Ours    & Matlab  & Chen et al.\cite{chen2021automatic}   \\ \hline
$f_x$  & 780.26  & 782.21  & 780.45  \\ \hline
$f_y$  & 1042.61 & 1046.03 & 1039.25 \\ \hline
$p_x$  & 745.43  & 752.34  & 761.34  \\ \hline
$p_y$  & 538.24  & 535.77  & 542.89  \\ \hline
RPE & 0.37    & 0.45    & 0.47    \\ \hline
STD & 0.02    & 0.10    & 0.24    \\ \hline
\end{tabular}
\end{table}

\textbf{Calibration on real data.}
To evaluate the performance of our system under realistic conditions, we perform calibration on a HIKROBOT MV-CA016-10GM camera with resolution of $1440 \times 1080$ by a $12 \times 8$ chessboard.
We repeat 20 times of calibration with different combinations of chessboard poses and get the average results of intrinsic parameters, RPE and the standard deviation(STD) of RPE.
The results compared with two state-of-art methods are provided in Table. \ref{table_rd}.
We can observe that the three systems produce similar results and ours gets the lowest RPE.
For better visualization, we draw the box-plot of three methods' RPE (Fig. \ref{fig_ipe_d}) in which we can seen the stability of each method and our RPE values are more stable than others.
Although Chen et al. \cite{chen2021automatic} can perform accurate detection, traditional parameter estimation's instability (with many abnormal value in Fig. \ref{fig_ipe_d}) reduces the calibration accuracy.

\subsection{Corner Detection Accuracy}
As calibration benefits from precise chessboard corner coordinates, the accuracy of our corner detection method is tested on synthetic data here.
Compared with both feature-based \cite{opencv_library,Geiger12} and learning-based \cite{kang2021sparse, chen2021automatic} corner detection methods, we conduct experiments on synthetic images with different configurations including extra noise, bad lighting and radial distortion ($k_0 = 1, k_1=-0.2, k_2=-0.1$).
Each configuration contains 2K chessboard images with ground truth sub-pixel corner coordinates.
The results are shown in Table. \ref{table_cl}.
It can be seen that our method gets the highest precision on different images, which is consistent with the calibration experiments and proves the precision of our method.

\begin{table}[t]
\centering
\small
\caption{Corner detection accuracy under different images}
\label{table_cl}
\setlength{\tabcolsep}{1.8mm}{
\begin{tabular}{|c||c|c|c|}
\hline
\multicolumn{4}{|c|}{Corner Detection Accuracy (pixels)} \\ \hline
\multirow{2}{*}{Methods}        & \multicolumn{3}{c|}{Image Configuration}         \\ \cline{2-4}
                                   &  Noise   & Bad Lighting & Distortion \\ \hline
OpenCV \cite{opencv_library} & 2.23 & 2.43 & 2.79 \\ \hline
libcb \cite{Geiger12} & 1.91 & 2.72 & 2.76 \\ \hline
Kang et al.\cite{kang2021sparse} & 1.25 & 1.58 & 1.72 \\ \hline
Chen et al. \cite{chen2021automatic} & 0.93 & 1.21 & 0.94 \\ \hline
Ours &  \textbf{0.78} & \textbf{0.51} & \textbf{0.71} \\ \hline

\end{tabular}}
\end{table}

\subsection{Distortion Correction performance}
To measure the contribution of our distortion correction part, we compare it with traditional optimization methods.
We combine both distortion correction methods with different detection and parameter estimation methods.
The experiments are conducted on our synthetic dataset whose distortion parameters are $k_0 = 1, k_1\!=\!-0.3, k_2 = -0.1$ and the average results are provided in Table \ref{Tab_cd}.
It can be seen that the distortion correction network improves the accuracy a lot in our framework.
Matlab combined with the distortion correction part gets accuracy improvement which is yet limited because the distortion correction brings into noise caused by image interpolation.
On the other hand, as the learning-based detection is robust against image noise, Chen et al. method gets precise result on images corrected by our distortion correction method.

\begin{table}[t]
\begin{threeparttable}[t]
\small
\caption{Calibration with/without Distortion correction}
\centering
\label{Tab_cd}
\setlength{\tabcolsep}{1.3mm}{
\begin{tabular}{|c|c|c|c|c|c|c|}
\hline
           & \multicolumn{2}{c|}{Ours} & \multicolumn{2}{c|}{Matlab}   & \multicolumn{2}{c|}{Chen et al.\cite{chen2021automatic}} \\ \hline
DC\tnote{1} &   $E_{IP}$   & $RPE$   & $E_{IP}$    & $RPE$   & $E_{IP}$   & $RPE$    \\ \hline
\small Optimization & 1.35    & 0.84    & 4.76   & 0.84    & 1.74    & 1.27    \\ \hline
\small Ours    & \textbf{0.79}    & \textbf{0.68}   & 4.41   & 0.96    & 0.93   & 0.71    \\ \hline
\end{tabular}}
\begin{tablenotes}
    \item[1] Different distortion correction methods.
\end{tablenotes}
\end{threeparttable}
\end{table}

\section{CONCLUSIONS}
In this paper, we propose a novel camera calibration framework containing three parts: distortion correction, corner detection and parameter estimation.
This framework integrates learning-based approaches with traditional methods to achieve efficient distortion correction, accurate sub-pixel feature detection and stable parameter estimation.
This framework surpasses other state-of-art calibration methods by a large margin in terms of accuracy on both synthetic and real dataset.
Additionally, extensive experiments prove the robustness of this framework against noise, lighting and radial distortion.
Besides, the corner detection and distortion correction part are evaluated respectively where decent results manifest the contribution of these parts to our framework. 











\bibliographystyle{IEEEtran}
\bibliography{ref}

\begin{thebibliography}{10}
\providecommand{\url}[1]{#1}
\csname url@rmstyle\endcsname
\providecommand{\newblock}{\relax}
\providecommand{\bibinfo}[2]{#2}
\providecommand\BIBentrySTDinterwordspacing{\spaceskip=0pt\relax}
\providecommand\BIBentryALTinterwordstretchfactor{4}
\providecommand\BIBentryALTinterwordspacing{\spaceskip=\fontdimen2\font plus
\BIBentryALTinterwordstretchfactor\fontdimen3\font minus
  \fontdimen4\font\relax}
\providecommand\BIBforeignlanguage[2]{{%
\expandafter\ifx\csname l@#1\endcsname\relax
\typeout{** WARNING: IEEEtran.bst: No hyphenation pattern has been}%
\typeout{** loaded for the language `#1'. Using the pattern for}%
\typeout{** the default language instead.}%
\else
\language=\csname l@#1\endcsname
\fi
#2}}

\bibitem{engel2017direct}
J.~Engel, V.~Koltun, and D.~Cremers, ``Direct sparse odometry,'' \emph{IEEE
  transactions on pattern analysis and machine intelligence}, vol.~40, no.~3,
  pp. 611--625, 2017.

\bibitem{SLAM:survey}
C.~Cadena, L.~Carlone, H.~Carrillo, Y.~Latif, D.~Scaramuzza, J.~Neira, I.~Reid,
  and J.~J. Leonard, ``Past, present, and future of simultaneous localization
  and mapping: Toward the robust-perception age,'' \emph{IEEE Transactions on
  Robotics}, vol.~32, no.~6, pp. 1309--1332, 2016.

\bibitem{CalibtoAD}
P.~F. Martins, H.~Costelha, L.~C. Bento, and C.~Neves, ``Monocular camera
  calibration for autonomous driving — a comparative study,'' in \emph{2020
  IEEE International Conference on Autonomous Robot Systems and Competitions
  (ICARSC)}, 2020, pp. 306--311.

\bibitem{barreto2009automatic}
J.~Barreto, J.~Roquette, P.~Sturm, and F.~Fonseca, ``Automatic camera
  calibration applied to medical endoscopy,'' in \emph{BMVC 2009-20th British
  Machine Vision Conference}.\hskip 1em plus 0.5em minus 0.4em\relax The
  British Machine Vision Association (BMVA), 2009, pp. 1--10.

\bibitem{opencv_library}
G.~Bradski, ``{The OpenCV Library},'' \emph{Dr. Dobb's Journal of Software
  Tools}, 2000.

\bibitem{MATLAB:2010}
MATLAB, \emph{version 7.10.0 (R2010a)}.\hskip 1em plus 0.5em minus 0.4em\relax
  Natick, Massachusetts: The MathWorks Inc., 2010.

\bibitem{Zhang:2000}
Z.~Zhang, ``A flexible new technique for camera calibration,'' \emph{Pattern
  Analysis and Machine Intelligence, IEEE Transactions on}, vol.~22, pp. 1330
  -- 1334, 12 2000.

\bibitem{Lucchese:2002}
L.~Lucchese and S.~Mitra, ``Using saddle points for subpixel feature detection
  in camera calibration targets,'' in \emph{Asia-Pacific Conference on Circuits
  and Systems}, vol.~2, 2002, pp. 191--195 vol.2.

\bibitem{SimonPlacht:2014ty}
S.~Placht, P.~F{\"u}rsattel, E.~A. Mengue, H.~Hofmann, C.~Schaller, M.~Balda,
  and E.~Angelopoulou, ``Rochade: Robust checkerboard advanced detection for
  camera calibration,'' in \emph{Computer Vision -- ECCV 2014}, D.~Fleet,
  T.~Pajdla, B.~Schiele, and T.~Tuytelaars, Eds.\hskip 1em plus 0.5em minus
  0.4em\relax Cham: Springer International Publishing, 2014, pp. 766--779.

\bibitem{Geiger12}
A.~Geiger, F.~Moosmann, O.~Car, and B.~Schuster, ``Automatic camera and range
  sensor calibration using a single shot,'' in \emph{International Conference
  on Robotics and Automation (ICRA)}, St. Paul, USA, May 2012.

\bibitem{Donne:2016eq}
S.~Donn{\'e}, J.~De~Vylder, B.~Goossens, and W.~Philips, ``{MATE: Machine
  Learning for Adaptive Calibration Template Detection},'' \emph{Sensors},
  vol.~16, no.~11, pp. 1858--17, Nov. 2016.

\bibitem{BenChen:2018fy}
B.~Chen, C.~Xiong, and Q.~Zhang, ``{CCDN: Checkerboard Corner Detection Network
  for Robust Camera Calibration},'' in \emph{Intelligent Robotics and
  Applications}.\hskip 1em plus 0.5em minus 0.4em\relax Cham: Springer, Cham,
  Aug. 2018, pp. 324--334.

\bibitem{Wu:2021}
H.~Wu and Y.~Wan, ``A highly accurate and robust deep checkerboard corner
  detector,'' \emph{Electronics Letters}, vol.~57, no.~8, pp. 317--320, 2021.

\bibitem{chen2021automatic}
B.~Chen, Y.~Liu, and C.~Xiong, ``Automatic checkerboard detection for robust
  camera calibration,'' in \emph{2021 IEEE International Conference on
  Multimedia and Expo (ICME)}.\hskip 1em plus 0.5em minus 0.4em\relax IEEE,
  2021, pp. 1--6.

\bibitem{Tang:2020bo}
Z.~Tang, R.~Grompone~von Gioi, P.~Monasse, and J.-M. Morel, ``{A Precision
  Analysis of Camera Distortion Models},'' \emph{IEEE Transactions on Image
  Processing}, vol.~26, no.~6, pp. 2694--2704, Sept. 2020.

\bibitem{salvi1998approach}
J.~Salvi \emph{et~al.}, \emph{An approach to coded structured light to obtain
  three dimensional information}.\hskip 1em plus 0.5em minus 0.4em\relax
  Universitat de Girona, 1998.

\bibitem{Tsai:1987}
R.~Tsai, ``A versatile camera calibration technique for high-accuracy 3d
  machine vision metrology using off-the-shelf tv cameras and lenses,''
  \emph{IEEE Journal on Robotics and Automation}, vol.~3, no.~4, pp. 323--344,
  1987.

\bibitem{Xue_2019_CVPR}
Z.~Xue, N.~Xue, G.-S. Xia, and W.~Shen, ``Learning to calibrate straight lines
  for fisheye image rectification,'' in \emph{Proceedings of the IEEE/CVF
  Conference on Computer Vision and Pattern Recognition (CVPR)}, June 2019.

\bibitem{Yin:2018tc}
X.~Yin, X.~Wang, J.~Yu, M.~Zhang, P.~Fua, and D.~Tao, ``Fisheyerecnet: A
  multi-context collaborative deep network for fisheye image rectification,''
  in \emph{Proceedings of the European conference on computer vision (ECCV)},
  2018, pp. 469--484.

\bibitem{Zhao:jp}
H.~Zhao, Y.~Shi, X.~Tong, and X.~Y. International, ``{A Simple Yet Effective
  Pipeline For Radial Distortion Correction},'' \emph{ieeexplore.ieee.org}, pp.
  878--882, 2020.

\bibitem{Sagawa:2008}
R.~Sagawa and Y.~Yagi, ``Accurate calibration of intrinsic camera parameters by
  observing parallel light pairs,'' 06 2008, pp. 1390 -- 1397.

\bibitem{GeoCalib}
J.-H. Chuang, C.-H. Ho, A.~Umam, H.-Y. Chen, J.-N. Hwang, and T.-A. Chen,
  ``Geometry-based camera calibration using closed-form solution of principal
  line,'' \emph{IEEE Transactions on Image Processing}, vol.~30, pp.
  2599--2610, 2021.

\bibitem{RANSAC}
\BIBentryALTinterwordspacing
M.~A. Fischler and R.~C. Bolles, ``Random sample consensus: A paradigm for
  model fitting with applications to image analysis and automated
  cartography,'' \emph{Commun. ACM}, vol.~24, no.~6, p. 381–395, June 1981.
  [Online]. Available: \url{https://doi.org/10.1145/358669.358692}
\BIBentrySTDinterwordspacing

\bibitem{ZHOU_ransac_points}
\BIBentryALTinterwordspacing
F.~Zhou, Y.~Cui, Y.~Wang, L.~Liu, and H.~Gao, ``Accurate and robust estimation
  of camera parameters using ransac,'' \emph{Optics and Lasers in Engineering},
  vol.~51, no.~3, pp. 197--212, 2013. [Online]. Available:
  \url{https://www.sciencedirect.com/science/article/pii/S0143816612003016}
\BIBentrySTDinterwordspacing

\bibitem{LV_ransac_pose}
\BIBentryALTinterwordspacing
Y.~Lv, J.~Feng, Z.~Li, W.~Liu, and J.~Cao, ``A new robust 2d camera calibration
  method using ransac,'' \emph{Optik}, vol. 126, no.~24, pp. 4910--4915, 2015.
  [Online]. Available:
  \url{https://www.sciencedirect.com/science/article/pii/S0030402615011869}
\BIBentrySTDinterwordspacing

\bibitem{schroeter2020learning}
J.~Schroeter, T.~Tuytelaars, K.~Sidorov, and D.~Marshall, ``Learning
  multi-instance sub-pixel point localization,'' in \emph{Proceedings of the
  Asian Conference on Computer Vision}, 2020.

\bibitem{kang2021sparse}
J.~Kang, H.~Yoon, S.~Lee, and S.~Lee, ``Sparse checkerboard corner detection
  from global perspective,'' in \emph{2021 IEEE International Conference on
  Signal and Image Processing Applications (ICSIPA)}.\hskip 1em plus 0.5em
  minus 0.4em\relax IEEE, 2021, pp. 12--17.

\bibitem{zhang2020distribution}
F.~Zhang, X.~Zhu, H.~Dai, M.~Ye, and C.~Zhu, ``Distribution-aware coordinate
  representation for human pose estimation,'' in \emph{Proceedings of the
  IEEE/CVF conference on computer vision and pattern recognition}, 2020, pp.
  7093--7102.

\bibitem{Rong:wf}
J.~Rong, S.~Huang, Z.~Shang, and X.~Ying, ``Radial lens distortion correction
  using convolutional neural networks trained with synthesized images,'' in
  \emph{ACCV}, 03 2017, pp. 35--49.

\bibitem{unet}
O.~Ronneberger, P.~Fischer, and T.~Brox, ``U-net: Convolutional networks for
  biomedical image segmentation,'' in \emph{International Conference on Medical
  image computing and computer-assisted intervention}.\hskip 1em plus 0.5em
  minus 0.4em\relax Springer, 2015, pp. 234--241.

\bibitem{TUMdata}
J.~Sturm, N.~Engelhard, F.~Endres, W.~Burgard, and D.~Cremers, ``A benchmark
  for the evaluation of rgb-d slam systems,'' in \emph{Proc. of the
  International Conference on Intelligent Robot Systems (IROS)}, Oct. 2012.

\bibitem{phong1975illumination}
B.~T. Phong, ``Illumination for computer generated pictures,''
  \emph{Communications of the ACM}, vol.~18, no.~6, pp. 311--317, 1975.

\end{thebibliography}

\end{document}